\documentclass{article}

\usepackage{arxiv}

\usepackage[utf8]{inputenc} 
\usepackage[T1]{fontenc}    
\usepackage{hyperref}       
\usepackage{url}            
\usepackage{booktabs}       
\usepackage{amsfonts}       
\usepackage{nicefrac}       
\usepackage{microtype}      
\usepackage{amsmath}
\usepackage{float} 
\usepackage{cleveref}       
\usepackage{lipsum}         
\usepackage{graphicx}
\usepackage{natbib}
\usepackage{doi}
\title{Measuring text summarization factuality using atomic facts entailment metrics in the context of retrieval augmented generation}

\newif\ifuniqueAffiliation
\uniqueAffiliationtrue

\ifuniqueAffiliation 
\author{ Nicolas Kriman F.\\
            Computational Data Analysis\\
            Georgia Institute of Technology
}
\else
\usepackage{authblk}

\newbox{\orcid}\sbox{\orcid}{\includegraphics[scale=0.06]{orcid.pdf}} 

\fi

\renewcommand{\shorttitle}{Measuring text summarization factuality }

\hypersetup{
pdftitle={Measuring text summarization factuality },
pdfsubject={q-bio.NC, q-bio.QM},
pdfauthor={Nicolas Kriman},
pdfkeywords={Retrieval Augmented Generation, Factuality Evaluation, Large Language Models},
}

\begin{document}
\maketitle

\begin{abstract}
Large language models (LLM) use has exploded since the introduction of ChatGPT in 2022, and it’s value has been evident in many use cases. One of its main barriers for enterprise/commercial use is its tendency to “make up” facts when answering, known as “hallucination”. In this project we propose a method for estimating the factuality of a summary given a source text using Naive Bayes classification. 
\end{abstract}


\section{Problem Statement}
Large language models (LLM) use has exploded since the introduction of ChatGPT in 2022, and it’s value has been evident in many use cases. One of its main barriers for enterprise/commercial use is its tendency to “make up” facts when answering, known as “hallucination”. There are several research topics that have tried to address this issue, and in this paper we will focus on measuring the factuality performance when using a technique called “Retrieval Augmented Generation” (RAG), which allows the LLM to use a reference document to generate an answer, assuming the reference document is correct. 

Using the following illustration we can understand how it works:
\begin{figure}[H]
    \centering
    \includegraphics[width=0.7\linewidth]{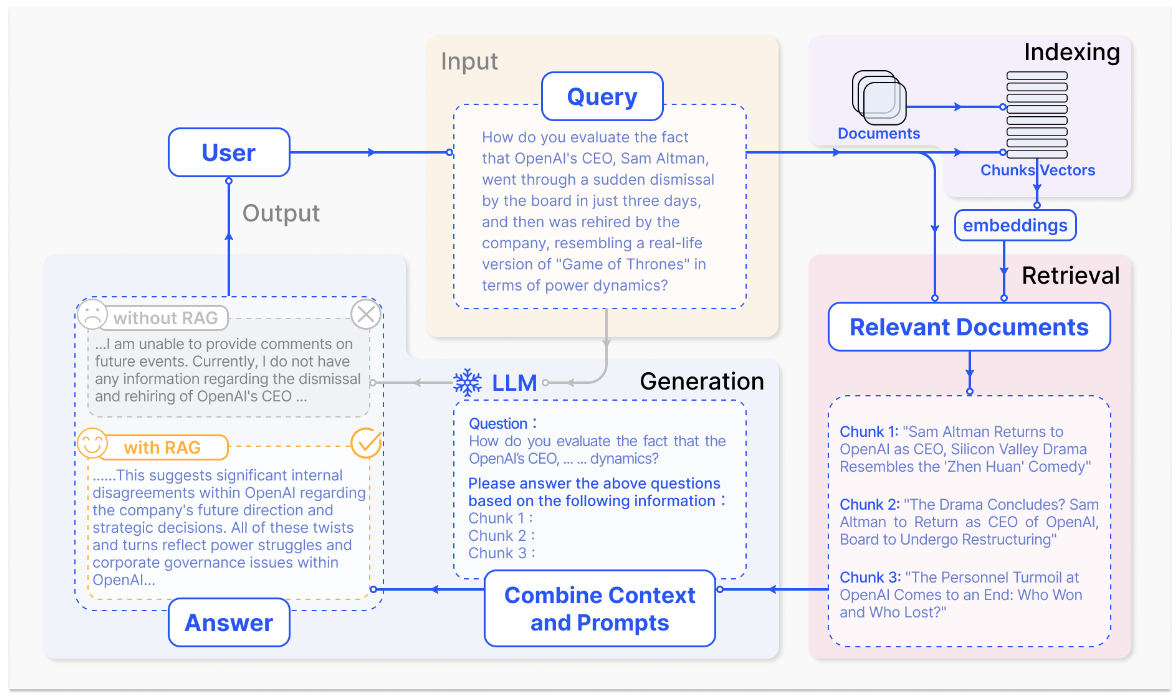}
    \caption{An overview of Retrieval Augmented Generation (Source \cite{Gao_Xiong_Gao_Jia_Pan_Bi_Dai_Sun_Wang_Wang_2024})
}
    \label{fig:fig1}
\end{figure}

\textbf{User Query Input:} The user inputs a query, such as ``How do you evaluate the fact that the previous CEO, Sam Altman, went through a sudden dismissal by the board in just three days, and then was rehired by the board within two days after the events were reviewed in terms of power dynamics?''

\textbf{Embedding Creation:} This query is transformed into a high-dimensional vector representation using an embedding algorithm. For example:
\begin{quote}
User query is transformed via embedding to Vector: \([-1, -0.23, 3, \ldots, 2.3]\)
\end{quote}

\textbf{Vector Search:} The generated query vector is then compared against a pre-existing set of vectors stored in a vector database using a specific search algorithm (e.g., BM25). 

\textbf{Search Results:} The search engine retrieves one or more vectors from the database that are deemed relevant to the input vector. 

\textbf{LLM Interpretation and Reasoning:} The large language model (LLM) interprets these retrieved vectors. It uses its ``reasoning capabilities'' to determine which of the retrieved vectors are most relevant to the query.

\textbf{Response Generation:} The LLM generates a final response based on the most relevant vectors and returns it to the user.

\textbf{Without RAG,} the answer could be: ``I am unable to provide comments on internal issues. Currently, I do not have enough information on the details and rationale of OpenAI's CEO turnover.'', while after the RAG processing the answer could be: ``This suggests significant internal disagreements with Sam Altman regarding the company's future direction and goals. The rapid turnaround indicates high-stakes power dynamics and potential governance issues within OpenAI.''

\section{Evaluation}
\subsection{RAG evaluation aspects}

We could evaluate different aspects of the RAG process:
\begin{itemize}
    \item Does it understand the underlying question? The user asks in natural language, so understanding the intent of the question could be not trivial. 
    \item Does it look for a direct (single-hop) answer or complex (multi-hop)? (See \cite{Mavi_Jangra_Jatowt_2024}): In broad terms, MHQA is the task of answering natural language questions that involve extracting and combining multiple pieces of information and doing multiple steps of reasoning.
    \item Once the “correct” question is embedded “perfectly”, how well does it retrieve the information from the vector database? Here we can think of how well the information is stored (e.g., granularity of the embedding: sentence, paragraph, document; embedding algorithm quality, etc.), how well the retrieval/search algorithm works, and most importantly how relevant the retrieved records are for answering the original question intent. This problem is well studied as a “search relevance” problem, and is similar to a search engine or a recommendation system.
    \item Once the “perfect” set of records is retrieved, and assuming they are coherent between them (but they could be incoherent and the final answer could state that explicitly), how well does the system filter (for relevance), combine (in a coherent way) and summarize them? This is the final “generation” step from “retrieval augmented generation”.
\end{itemize}
My evaluation is specific for the summarization step, which is an old NLP topic but with renewed relevance and new tools at hand. The question is: given a perfectly relevant document, is the summarization faithful / factual? We will be focusing on unlabeled content and for closed domain, meaning that we assume that the answer is completely contained in the retrieved document (open domain QA is a harder problem, as we would need to establish a massive ground truth knowledge base for my proposed methodology). This question is extremely relevant now that RAG is being implemented for enterprise level applications, where factuality and faithfulness to the enterprise knowledge base is crucial.
\subsection{Context for summarization performance measurement}
Traditional summary measurement uses a ROUGE score. This basically measures the overlap of exact words between the source text and its summary, with some variations (e.g., varying n-grams, etc). This is mostly useless for capturing the semantic consistency. We also have other classic scores as BLEU which was originally designed for machine translation, it calculates precision of n-grams in the candidate text relative to a reference text. Incorporates brevity penalty to avoid overly short translations or summaries. Focuses on precision and exact match of n-grams, which may not reflect the true semantic equivalence between the texts. We also have METEOR which is similar to BLEU with some more sophistication to consider synonyms for example, but has very similar problems. The main issue with BLEU and METEOR is that they were created for evaluation of “Machine translation”, so they rely on having a reference translation to compare to. This is not useful for our use case, as we are trying to measure unlabeled summarizations.

A big step forward was made with the introduction of BERT, which in turn allowed for the calculation of a BERTscore. This score was calculated by embedding the source sentence and its summary (or translation) and then by measuring their similarity using cosine distance. But this has the same problems that we, many years later, still have with LLMs: they struggle with semantic alignment. Two sentences that have opposite meanings can have very similar scores. We can see in practice the performance of these traditional metrics in this table:
\begin{figure}[H]
    \centering
    \includegraphics[width=0.7\linewidth]{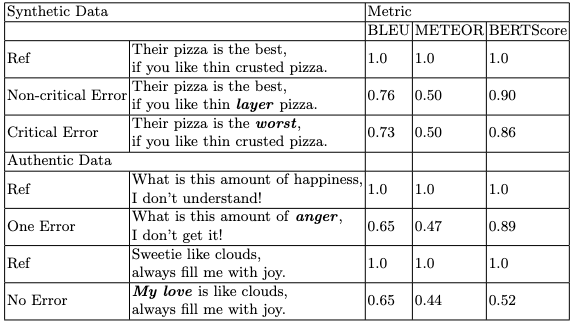}
    \caption{A review of traditional summarization metrics (Source: \cite{Saadany_Orasan_2021}
)
}
    \label{fig:fig2}
\end{figure}

It was already discussed in \cite{Fabbri_Wu_Liu_Xiong_2022} that all traditional metrics are poor evaluators of factuality.

My proposed method uses yet another methodology, which is similar to recent advances (\cite{Muhlgay_Ram_Magar_Levine_Ratner_Belinkov_Abend_Leyton-Brown_Shashua_Shoham_2024} and \cite{Min_Krishna_Lyu_Lewis_Yih_Koh_Iyyer_Zettlemoyer_Hajishirzi_2023}). In summary, the methodology is using traditional LLM generation (with some variations in what is being measured), and the metric relies on the next token probability for classification.

\section{Data Source}
We are going to evaluate this using the AggreFact benchmark and compare against other algorithms. Here we have a description of each field of the dataset:
\begin{table}[H]
\centering
\begin{tabular}{|l|p{10cm}|}
\hline
\textbf{Col. name}    & \textbf{Description}                                                                                                 \\ \hline
dataset               & Name of the original annotated dataset.                                                                             \\ \hline
origin                & Summarization dataset. Either cnndm or xsum.                                                                        \\ \hline
id                    & Document id.                                                                                                         \\ \hline
doc                   & Input article.                                                                                                       \\ \hline
summary               & Model generated summary.                                                                                            \\ \hline
model name            & Name of the model used to generate the summary.                                                                     \\ \hline
label                 & Factual consistency of the generated summary. 1 is factually consistent, 0 otherwise.                               \\ \hline
cut                   & Either val or test.                                                                                                  \\ \hline
system score          & The output score from a factuality system.                                                                           \\ \hline
system label          & The binary factual consistency label based on the score of the factuality system. Only examples in the test set have labels. Labels are determined under the threshold-per-dataset setting. \\ \hline
\end{tabular}
\caption{Description of the table columns}
\label{tab:col_description}
\end{table}

And here we can see a single sample:

\begin{table}[H]
\centering
\begin{tabular}{|l|p{10cm}|}
\hline
\textbf{Field} & \textbf{Content} \\ \hline
dataset & XSumFaith \\ \hline
origin & xsum \\ \hline
id & 34687720 \\ \hline
doc & France's Dubuisson carded a 67 to tie with overnight leader Van Zyl of South Africa on 16 under par. McIlroy carded a third straight five under-par 67 to move to 15 under par with Thailand's Kiradech Aphibarnrat. The world number three's round included an eagle on the 12th as he bids to win his first title since May. "The 67s I've shot this week have all been a little different and I feel like I've played within myself for all of them, " said four-time major winner McIlroy of Northern Ireland. "I feel there's a low [...] \\ \hline
summary & rory mcilroy will take a one-shot lead into the final round of the wgc-hsbc champions after carding a three-under \\ \hline
model name & BERTS2S \\ \hline
label & 0 \\ \hline
cut & val \\ \hline
DAE score & 0.00841161 \\ \hline
DAE label &  \\ \hline
QuestEval score & 0.35180809121165 \\ \hline
QuestEval label &  \\ \hline
SummaC-ZS score & -0.1430435180664062 \\ \hline
SummaC-ZS label &  \\ \hline
SummaC-Conv score & 0.2148666381835936 \\ \hline
SummaC-Conv label &  \\ \hline
QAFactEval score & 0.0 \\ \hline
QAFactEval label &  \\ \hline
\end{tabular}
\caption{Data sample}
\label{tab:data_sample}
\end{table}

Given the massive dataset size and the high cost of either using APIs or using GPU intensive resources, I will sample the dataset to obtain 90\% (+/- 10\%) confidence interval, which results in around 70 samples. The dataset is in the repository \\
\url{https://github.com/Liyan06/AggreFact/}

\section{Methodology}

We are going to try to reproduce the methodology and results (comparing to human processing) described in \cite{Min_Krishna_Lyu_Lewis_Yih_Koh_Iyyer_Zettlemoyer_Hajishirzi_2023}, combined with the factuality taxonomies from \cite{Pagnoni_Balachandran_Tsvetkov_2021a}.

First we fine-tune an LLM model to disaggregate a source text into several “atomic facts” (pieces of information).

With the model trained we can now disaggregate both the source text and the summary, for example:\\
Source text: “The Federal Aviation Administration on Wednesday initially reported a pressurization problem with SkyWest Flight 5622, and said it would investigate. It later issued a statement that did not reference any pressurization issues.” \\
Source text atomic facts:\\
\begin{itemize}
    \item The Federal Aviation Administration reported a pressurization problem with SkyWest Flight 5622.
    \item The FAA reported the pressurization problem on Wednesday.
    \item The FAA said it would investigate.
    \item The FAA later issued a statement that did not reference any pressurization issues.
\end{itemize}
Summary text: “The airline says it's investigating the cause of a pressurization problem”
Summary text atomic facts:
\begin{itemize}
    \item The airline says it is investigating the cause.
    \item The airline is investigating a pressurization problem.
\end{itemize}

\section{Approach}
In this section, we describe our approach for evaluating the factuality of summary texts with respect to their source texts. Our methodology leverages a large language model (LLM) to cross-compare atomic facts from the summary and source, using the factuality categories from \cite{Pagnoni_Balachandran_Tsvetkov_2021a}. After estimating the probability for each category, we train a Naive Bayes classifier to finally predict if the summarization is factual or not.

\subsection{Atomic Fact Extraction}
The first step is we extract atomic facts from both the source and summary texts. Atomic facts are the smallest units of factual information that can stand alone and be evaluated independently. These extractions are performed using a LLM.

\subsection{Factuality Categories}
Each atomic fact from the summary is compared with the atomic facts from the source using the following factuality categories:
\begin{itemize}
    \item \textbf{PredE (Relation Error)}: The predicate in the summary statement is inconsistent with the source article.
    \item \textbf{EntE (Entity Error)}: The primary arguments (or their attributes) of the predicate are incorrect.
    \item \textbf{CircE (Circumstance Error)}: Additional information (like location or time) specifying the circumstance around a predicate is incorrect.
    \item \textbf{CorefE (Coreference Error)}: A pronoun or reference has an incorrect or non-existent antecedent.
    \item \textbf{LinkE (Discourse Link Error)}: Error in how multiple statements are linked together in the discourse (e.g., temporal ordering or causal link).
    \item \textbf{OutE (Out of Article Error)}: The statement contains information not present in the source article.
    \item \textbf{GramE (Grammatical Error)}: The grammar of the sentence is so erroneous that it becomes meaningless.
    \item \textbf{Perfectly supported}: The atomic fact in the summary is perfectly supported by the atomic fact in the source text (this category was created for this project and not present in the original paper).
\end{itemize}

\subsection{Cross-Comparison Using LLM}
Now both the source text and summary text have their own set of atomic facts.
For each combination of summary atomic fact and source atomic fact, we input them into a fine-tuned LLM to perform a cross-comparison. The model is fine-tuned to predict the likelihood of encountering each of the factuality categories based on the input pairs.
For example, the prompt could be "Given the source text $source_i$ and the summary text $summary_i$,  the predicate in the summary statement is inconsistent with the source article?" and we can observe that the next token probabilities are:"$yes = 0.9$, $no = 0.1$".
We would expect a perfectly factual summary text to have all negative features equal to zero, and the positive feature equal to one. 

\subsection{Factuality Measurement}
Given that we now have 7 negative features and 1 positive feature, each with a value between 0 and 1, and a target classification "factual / not factual", we arrive at a classic classification problem. For this project I will use a Naive Bayes classifier
\subsection{Naive Bayes Classifier}
The Naive Bayes classifier is based on Bayes' theorem, which provides a way to calculate the probability of a hypothesis given prior knowledge. In our context, the hypothesis is whether a summary atomic fact is "factual" or "not factual" given the observed features.

Bayes' theorem is given by:

\[
P(C \mid \mathbf{X}) = \frac{P(\mathbf{X} \mid C) P(C)}{P(\mathbf{X})}
\]

where:\\
- \( P(C \mid \mathbf{X}) \) is the posterior probability of class \( C \) (factual or not factual) given the feature vector \( \mathbf{X} \).\\
- \( P(\mathbf{X} \mid C) \) is the likelihood of the feature vector given the class.\\
- \( P(C) \) is the prior probability of the class.\\
- \( P(\mathbf{X}) \) is the prior probability of the feature vector.\\

Since we are using a Naive Bayes classifier, we assume that the features are conditionally independent given the class. This simplifies the calculation of the likelihood:

\[
P(\mathbf{X} \mid C) = \prod_{i=1}^{n} P(X_i \mid C)
\]

where \( X_i \) is the \( i \)-th feature in the feature vector \( \mathbf{X} \), and \( n \) is the total number of features (in our case, \( n = 7 \)).

\subsection{Feature Probabilities}
Each feature \( X_i \) represents the probability of a specific factuality category being applicable to the given atomic fact pair. 

\subsection{Posterior Probability Calculation}
The posterior probability for the class "factual" can be written as:

\[
P(\text{factual} \mid \mathbf{X}) = \frac{P(\mathbf{X} \mid \text{factual}) P(\text{factual})}{P(\mathbf{X})}
\]

Similarly, the posterior probability for the class "not factual" can be written as:

\[
P(\text{not factual} \mid \mathbf{X}) = \frac{P(\mathbf{X} \mid \text{not factual}) P(\text{not factual})}{P(\mathbf{X})}
\]

Since \( P(\mathbf{X}) \) is constant for both classes, it can be ignored in the final classification decision. Thus, the decision rule is to choose the class that maximizes the posterior probability:

\[
\hat{C} = \arg\max_{C \in \{\text{factual}, \text{not factual}\}} P(C \mid \mathbf{X})
\]

\subsection{Implementation}
Given the feature independence assumption, the decision rule can be simplified as:

\[
\hat{C} = \arg\max_{C \in \{\text{factual}, \text{not factual}\}} P(C) \prod_{i=1}^{n} P(X_i \mid C)
\]

This approach allows us to classify each summary atomic fact as either factual or not factual based on the observed features, leveraging the probabilities provided by the fine-tuned LLM.

For the scope of this project, we will judge the entire summary by its worse performing atomic fact.

\section{Implementation}
The original implementation plan was to separate both source and summary texts into atomic facts and focus more on the evaluation. In practice, the evaluation probed to be relatively easy, and the separation of the original texts into atomic facts was difficult. I decided that given this project was focused on the Naive Bayes implementation, I would manually identify atomic facts. This is a topic for further development of this technique. Identification of facts in a text is a hard problem, given that for example one fact must be deduced from multiple sentences. 
Given the separated atomic facts from the original texts, it was easy to use an LLM to evaluate a source-summary pair for factuality using the factuality categories. After that, each pair evaluation used next-token logit probabilities given by a LLM, as described in section 5.3 .

\section{Evaluation}

We can see the different metrics evaluation against the benchmarks in the following table:

\begin{table}[h!]
\centering
\begin{tabular}{|l|c|c|c|c|}
\hline
\textbf{Model} & \textbf{AUC} & \textbf{Accuracy} & \textbf{F1} & \textbf{Precision} \\
\hline
DAE\_label & 0.65 & 0.65 & 0.67 & 0.63 \\
SummaC-ZS\_label & 0.70 & 0.70 & 0.71 & 0.68 \\
SummaC-Conv\_label & 0.73 & 0.73 & 0.75 & 0.70 \\
QAFactEval\_label & 0.65 & 0.65 & 0.68 & 0.63 \\
gpt-3.5-turbo & 0.75 & 0.75 & 0.77 & 0.70 \\
Naive Bayes (*) & 0.53 & 0.52 & 0.46 & 0.60 \\
\hline
\end{tabular}
\caption{Performance metrics for different models}
\label{table:1}
\end{table}

Now the interesting part is: ¿why did this model perform poorly?

I've identified several issues that could explain its performance and be improved in a future implementation.

\subsection{Factuality categories correlation}

In the original paper \cite{Pagnoni_Balachandran_Tsvetkov_2021a} , the evaluation of factuality was done by humans (who even had some disagreement on how to evaluate the categories in real text). The evaluation done in this work was done using an LLM, an algorithm that is very sensible to its training data, and the "prompt" (the exact wording of the input text). If we do a simple correlation analysis we can see that all factuality categories evaluation are highly correlated.

\begin{figure}[H]
    \centering
    \includegraphics[width=0.75\linewidth]{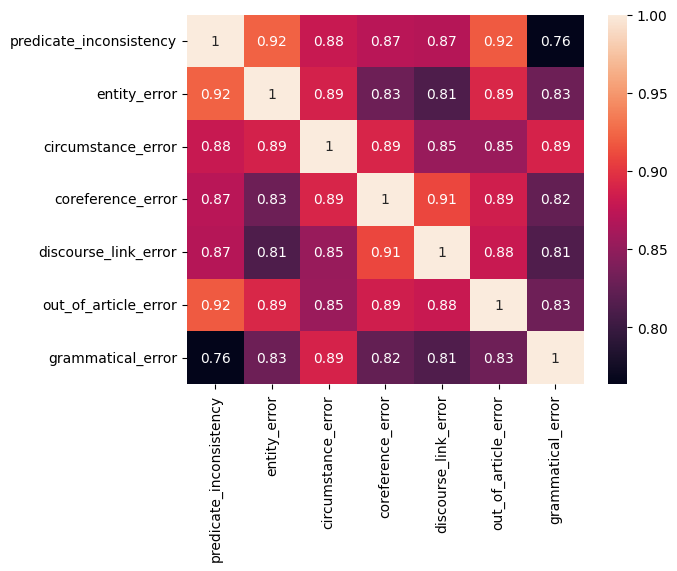}
    \caption{Factuality categories evaluation correlation}
    \label{fig:enter-label}
\end{figure}

\subsection{No clear separation of categories using PCA}
If we do a PCA reduction, using just 2 dimensions captures 92\% of the variance. When visualizing the reduced dimension plot, we can not see a clear separation between factual (1) and non-factual (0) summaries, even if trying to separate them using a non-linear kernel (both categories are intertwined), which could impact several other models performance:

\begin{figure}[H]
    \centering
    \includegraphics[width=0.75\linewidth]{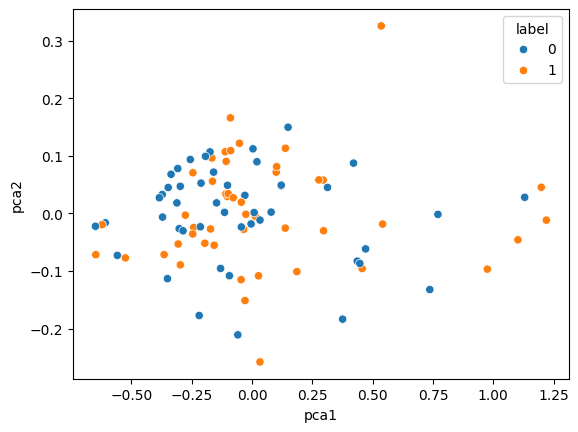}
    \caption{PCA representation of factuality categories evaluation}
    \label{fig:enter-label}
\end{figure}

\subsection{Multi-hop logic is required for a correct evaluation}

This is one of the most difficult tasks required to do this correctly, and I think it is one of the main reasons of its poor performance. Let's consider the follow source text atomic facts:
\begin{itemize}
    \item  'Barrow winger Williams scored a fine 48th-minute opener for Barrow.
    \item  'Ryan Yates scored the winning goal for Barrow.
    \item  'Barrow won the match 2-1 against Taunton Town.'
\end{itemize}
And the summary text atomic fact to be evaluated is 'Jordan Williams scored twice as Barrow beat League Two side Taunton.'

The summary states that player "Jordan Williams" scored two goals against the team called "Taunton". If we compare the summary atomic fact against each source text atomic fact independently, we can see that there should be no error of factuality. If the scored a goal at the 48th minute, it is factual. If "Ryan Yates" scored a goal, it is not contrary to the summary statement. And if "Barrow" won 2-1 against "Taunton", it actually supports the summary statement. It is only when we evaluate the entire corpus of atomic facts that we realize that the summary statement is wrong. 

\section{Future directions}

Given the challenges identified in the implementation and evaluation of the Naive Bayes model for factuality assessment, several future directions could enhance the model's performance. 

\subsection{Named Entity Recognition and Entity Disambiguation}

To address the challenges in identifying atomic facts, integrating NER can help in recognizing and categorizing entities such as names, dates, and locations within the text, while entity disambiguation can resolve ambiguities when multiple entities share the same name (or variations of it). 

Named Entity Recognition (NER) is the process of identifying and classifying named entities mentioned in text into predefined categories such as the names of people, organizations, locations, dates, and other proper nouns. NER systems scan text to detect entities, typically using a combination of linguistic grammar-based techniques and machine learning models trained on annotated datasets. Once entities are identified, they are classified into specific categories. For example, "Barack Obama" would be classified as a person, "Microsoft" as an organization, and "New York" as a location.

NER systems leverage advanced algorithms and large annotated corpora to achieve high accuracy. However, challenges remain, particularly in dealing with entities that have multiple meanings or those that are context-dependent.

Entity disambiguation, also known as named entity disambiguation or entity linking, is the process of resolving ambiguities when an entity mentioned in the text could refer to multiple possible real-world entities (or entities that are present in the source text, for this paper application). This technique is needed for ensuring that the correct entity is identified. Disambiguation algorithms consider the context surrounding the entity mention to determine the most likely real-world entity. For example, the name "Apple" could refer to the fruit or the technology company, and context is used to distinguish between these meanings.

The practical application of NER and Entity Disambiguation for this case is that we can identify entities that are common between the source and summary. A good summary should retain all important entities present in the source text. Using our previous example, the source text entities should be:

\textbf{Source text:}
\begin{quote}
    Barrow winger Williams scored a fine 48th-minute opener for Barrow. Ryan Yates scored the winning goal for Barrow. Barrow won the match 2-1 against Taunton Town.
\end{quote}

\textbf{Identified Entities:}
\begin{itemize}
    \item \textbf{Person:}
    \begin{itemize}
        \item Williams
        \item Ryan Yates
    \end{itemize}
    \item \textbf{Organization/Team:}
    \begin{itemize}
        \item Barrow
        \item Taunton Town
    \end{itemize}
    \item \textbf{Event:}
    \begin{itemize}
        \item 48th-minute opener
    \end{itemize}
    \item \textbf{Score:}
    \begin{itemize}
        \item 2-1
    \end{itemize}
\end{itemize}

\textbf{Summary text:}
\begin{quote}
    Jordan Williams scored twice as Barrow beat League Two side Taunton.
\end{quote}

\textbf{Identified Entities:}
\begin{itemize}
    \item \textbf{Person:}
    \begin{itemize}
        \item Jordan Williams
    \end{itemize}
    \item \textbf{Organization/Team:}
    \begin{itemize}
        \item Barrow
        \item Taunton (assumed to be Taunton Town)
    \end{itemize}
\end{itemize}

\textbf{Disambiguation:}
\begin{itemize}
    \item \textbf{Person:}
    \begin{itemize}
        \item ``Williams'' in the source is the same as ``Jordan Williams'' in the summary.
        \item ``Ryan Yates'' is missing from the summary.
    \end{itemize}
    \item \textbf{Organization/Team:}
    \begin{itemize}
        \item ``Barrow'' is present in both source and summary.
        \item ``Taunton'' in the summary refers to ``Taunton Town'' in the source.
    \end{itemize}
    \item \textbf{Event:}
    \begin{itemize}
        \item The specific ``48th-minute opener'' is not mentioned in the summary.
    \end{itemize}
    \item \textbf{Score:}
    \begin{itemize}
        \item The specific match score ``2-1'' is not mentioned in the summary.
    \end{itemize}
\end{itemize}

\subsection{Multi-hop QA Reasoning}

Multi-hop QA reasoning involves answering questions that require integrating information from multiple parts of a text. Implementing multi-hop QA reasoning can help the model understand and evaluate summaries that require synthesizing information from multiple sources.

\subsection{Pre-trained Entailment Models for Atomic Facts Comparison}

Utilizing pre-trained entailment models can be simpler and more effective than fine-tuning a LLM (which is focused on text generation instead of text classification). These models are trained on large corpora to understand entailment relationships, which can be directly applied to evaluate whether a summary factually aligns with the source text.

\nocite{*}
\bibliography{references}

\end{document}